\documentclass[journal]{IEEEtran}
\usepackage{amsmath,amsfonts}
\usepackage{algorithmic}
\usepackage{algorithm}
\usepackage{array}
\usepackage[caption=false,font=normalsize,labelfont=sf,textfont=sf]{subfig}
\usepackage{textcomp}
\usepackage{stfloats}
\usepackage{url}
\usepackage{orcidlink}
\usepackage{orcidlink}
\usepackage{tikz}
\usepackage{verbatim}
\usepackage{graphicx}
\usepackage{multirow}
\usepackage{booktabs}
\usepackage{pifont}
\newcommand{\cmark}{\ding{51}}
\newcommand{\xmark}{\ding{55}}
\usepackage[table]{xcolor}
\usepackage{biblatex}
\usepackage[T1]{fontenc}
\def\BibTeX{{\rm B\kern-.05em{\sc i\kern-.025em b}\kern-.08em T\kern-.1667em\lower.7ex\hbox{E}\kern-.125emX}}
\usepackage{balance}
\hypersetup{
colorlinks=true,
linkcolor=black,
citecolor=black
}

\begin{document}
\title{Human4K: A Large-Scale 4K Multi-View Mocap Dataset for Whole-Body 3D Human Reconstruction}
\author{Tianshun Han~\orcidlink{0009-0004-3393-1597}, 
Ziyu~Shi,
Lijian Liu, 
Ajian Liu~\orcidlink{0000-0002-7788-9368},
Benjia Zhou~\orcidlink{0000-0003-4883-5552},
Hugo~Jair~Escalante~\orcidlink{0000-0003-4603-3513}, \\
Yanyan Liang$^{*}$~\orcidlink{0000-0002-5780-8540},~\IEEEmembership{Member,~IEEE}, 
Sergio~Escalera~\orcidlink{0000-0003-0617-8873},~\IEEEmembership{Senior~Member,~IEEE}, \\
Zhen~Lei~\orcidlink{0000-0002-0791-189X},~\IEEEmembership{Fellow,~IEEE}, 
Jun Wan$^{*}$~\orcidlink{0000-0002-4735-2885},~\IEEEmembership{Senior Member,~IEEE}
\thanks{
Tianshun Han and Yanyan Liang are with the School of Computer Science and Engineering, the Faculty of Innovation Engineering, Macau University of Science and Technology, Macau 999078, China. (e-mail: \{3230002542\}@student.must.edu.mo, \{yyliang\}@must.edu.mo).

Lijian Liu, Ziyu Shi, Ajian Liu and Jun Wan are with the State Key Laboratory of Multimodal Artificial Intelligence Systems (MAIS), Institute of Automation, Chinese Academy of Sciences (CASIA), Beijing 100190, China, also with the School of Artificial Intelligence, University of Chinese Academy of Sciences (UCAS), Beijing 100049, China. Ajian Liu and Jun Wan are also with the School of Computer Science and Engineering, the Faculty of Innovation Engineering, Macau University of Science and Technology, Macau 999078, China. (e-mail: \{lijian.liu, ziyu.shi, ajian.liu, jun.wan\}@ia.ac.cn).

Benjia Zhou is with Beijing Institute of Technology, Zhuhai, Zhuhai 519088, China. (e-mail: \{benjiazhou\}@bit.edu.cn).

Hugo~Jair~Escalante is with Instituto Nacional de Astrofísica, Óptica y Electrónica, Puebla, Mexico, and also with Computer Science Department, CINVESTAV Zacatenco, Mexico City 07360, Mexico. (e-mail: \{hugojair\}@inaoep.mx).

Sergio~Escalera is with the Computer Vision Center, 08193 Barcelona, Spain, also with Universitat de Barcelona, 08007 Barcelona, Spain, and also with Aalborg University, 9220 Aalborg, Denmark. (e-mail: \{sergio\}@maia.ub.es).

Zhen Lei is with the State Key Laboratory of Multimodal Artificial Intelligence Systems (MAIS), Institute of Automation Chinese Academy of Sciences (CASIA),Beijing 100190, China, also with the School of Artificial Intelligence, University of Chinese Academy of Sciences (UCAS), Beijing 100049, China, and also with the Centre for Artificial Intelligence and Robotics, Hong Kong Institute of Science and Innovation, Chinese Academy of Sciences, Hong Kong, China (e-mail: \{zhen.lei@ia.ac.cn\}).

\emph{$^{*}$Co-corresponding authors: Yanyan~Liang and Jun~Wan.}

}}

\markboth{}{Human4K: A Large-Scale 4K Multi-View Mocap Dataset for Whole-Body 3D Human Reconstruction}
\maketitle 
\begin{abstract}
Recent advances in 3D human reconstruction have improved overall performance, yet current models still fail in the most challenging real-world scenarios. They often produce unstable geometry, inaccurate limb articulation and unreliable predictions under depth ambiguity or self-occlusion. A key reason is that existing datasets still lack the combination of high-resolution images, high-precision annotations and diverse whole-body motions required to support robust reconstruction. To address this gap, we present Human4K, a large-scale 4K multi-view whole-body human reconstruction dataset with mocap-accurate SMPL-X annotations. Human4K contains over six million 4K images captured by an eight-view high-resolution camera system synchronized with a professional Vicon motion capture setup, covering 11 subjects performing complex, highly articulated and strongly self-occluded full-body motions. All sequences are processed by a Motion-Retargeting and Refinement Module (MRRM) to ensure precise alignment for the full body and extremities. Experimental results show that training with Human4K consistently improves whole-body reconstruction on standard benchmarks, with particularly large gains for hands, feet and depth-ambiguous limb configurations. \footnote{Human4K will be released when this paper is accepted.}
\end{abstract}

\begin{table}[!t]
\setlength{\tabcolsep}{5pt} % 缩小列间距
\renewcommand{\arraystretch}{1.2}
\caption{\textbf{Comparison of existing datasets and Human4K.} Human4K is a large-scale 4K full-body mocap dataset with accurate SMPL-X annotations and richer whole-body motions than prior datasets.
\textbf{Frames}: number of annotated frames. 
\textbf{Video}: whether video sequences are provided. 
\textbf{Cx}: whether complex motions are included. 
\textbf{Res}: image resolution. 
\textbf{Skel}: annotated skeleton type. 
\textbf{Subj}: number of subjects.}
\begin{tabular}{@{}lcccccc@{}}
\toprule
\textbf{Dataset} & \textbf{\#Frames} & \textbf{Video} & \textbf{Cx} & \textbf{Res} & \textbf{Skel} & \textbf{Subj} \\
\midrule

\rowcolor{gray!15}
\multicolumn{7}{c}{\textbf{Pseudo-3D}} \\

MSCOCO~\cite{lin2014microsoft}       & 200K   & \cmark & \xmark & 1K & body    & -- \\
MPII~\cite{andriluka20142d}          & 25K    & \xmark & \xmark & 1K & body    & -- \\
LSPET~\cite{johnson2010clustered}    & 2.9K   & \xmark & \xmark & 1K & body    & -- \\
OCHuman~\cite{zhang2019pose2seg}     & 2.5K   & \xmark & \xmark & 1K & body    & -- \\
Ubody~\cite{lin2023one}              & 683.3K & \cmark & \xmark & 1K & body    & -- \\
\midrule

\rowcolor{gray!15}
\multicolumn{7}{c}{\textbf{Synthetic}} \\

AGORA~\cite{Patel:CVPR:2021}          & 17K    & \xmark & \xmark & 1K & W-body & $>$350 \\
Synbody~\cite{yang2023synbody}       & 633.5K & \cmark & \xmark & 1K & W-body & 10000 \\
BEDLAM~\cite{black2023bedlam}        & 951.1K & \xmark & \xmark & 1K & W-body & 271 \\
GTA-human~\cite{cai2021playing}      & 1.4M   & \cmark & \xmark & 1K & body   & $>$600 \\
SURREAL~\cite{varol17_surreal}       & 6.5M   & \cmark & \xmark & 1K & body   & 145 \\
\midrule

\rowcolor{gray!15}
\multicolumn{7}{c}{\textbf{Multiview}} \\

MPI\_INF\_3DHP~\cite{mono-3dhp2017}    & 1.3M   & \cmark & \cmark & 1K & body   & 8 \\
ZJU-Mocap~\cite{easymocap}            & 237K   & \cmark & \xmark & 1K & body   & 9 \\
DNA-rendering~\cite{cheng2023dna}     & 67.5M  & \cmark & \cmark & 4K & W-body & 1500 \\
MVhumanNet~\cite{xiong2024mvhumannet} & 645.1M & \cmark & \cmark & 4K & W-body & 4500 \\
HuMMan~\cite{cai2022humman}           & 40M    & \xmark & \xmark & 1K & body   & 100 \\
Genebody~\cite{cheng2022generalizable}& 360M   & \xmark & \xmark & 4K & W-body & 370 \\
FreeMan~\cite{wang2024freeman}        & 11.3M  & \xmark & \xmark & 1K & body   & 40 \\
\midrule

\rowcolor{gray!15}
\multicolumn{7}{c}{\textbf{Mocap}} \\

EHF~\cite{SMPL-X:2019}                & 100    & \xmark & \xmark & 1K & --      & 1 \\
3DPW~\cite{vonMarcard2018}            & 51K    & \cmark & \xmark & 1K & body    & 18 \\
Human3.6M~\cite{6682899}              & 3.6M   & \xmark & \xmark & 1K & body    & 11 \\
\textbf{Human4K (Ours)}               & \textbf{6M} & \cmark & \cmark & \textbf{4K} & \textbf{W-body} & \textbf{11} \\
\bottomrule
\end{tabular}
\label{table1}
\end{table}

\section{Introduction}
In recent years, significant advancements have been made in estimating human posture and shape from monocular images or videos, playing a crucial role in enabling applications in augmented reality (AR) and virtual reality (VR). Current reconstruction pipelines typically rely on parametric human models to recover body shape and pose, and require large datasets that provide accurate geometric supervision together with diverse whole-body motions. Therefore, over the past decade, a variety of datasets have been developed, ranging from pseudo-3D annotated datasets~\cite{lin2014microsoft, andriluka20142d, johnson2010clustered, zhang2019pose2seg, lin2023one} and synthetic datasets~\cite{yang2023synbody, Patel:CVPR:2021, black2023bedlam, cai2021playing, varol17_surreal} to multiview and motion-capture-based datasets~\cite{6682899, SMPL-X:2019, vonMarcard2018}.

However, despite these advances, existing datasets still exhibit fundamental limitations (summarized in Table~\ref{table1}) that hinder the development of high-fidelity 3D human reconstruction models for real-world scenarios. These datasets can be roughly grouped into four categories.
\textbf{(1) Pseudo-3D datasets}~\cite{andriluka20142d, lin2014microsoft, johnson2010clustered, zhang2019pose2seg, lin2023one} lift 2D keypoints to SMPL or SMPL-X parameters via annotators such as NeuralAnnot~\cite{Moon_2022_CVPRW_NeuralAnnot}, which makes them easy to scale but noisy and coarse for extremities such as ankles, wrists and fingers. \textbf{(2) Synthetic datasets}~\cite{varol17_surreal, Patel:CVPR:2021, yang2023synbody, black2023bedlam, cai2021playing} provide accurate and scalable annotations, but the domain gap between synthetic and real imagery causes poor generalization, severely limiting their practical applicability. \textbf{(3) Multiview datasets}~\cite{mono-3dhp2017,easymocap,cheng2023dna,xiong2024mvhumannet,cai2022humman} capture large-scale real data with multiple cameras, but still rely on multiview optimization or pseudo labels that are fragile under severe occlusion or extreme articulation. \textbf{(4) Mocap datasets}~\cite{vonMarcard2018,6682899,sigal2006humaneva,AMASS:ICCV:2019,SMPL-X:2019,kaufmann2023emdb} alleviate some of these issues by providing high-quality 3D keypoints from marker-based systems, making them more reliable than pseudo- or multiview-based annotations. However, traditional mocap data typically lacks high-resolution imagery, offers limited diversity of complex or extreme motions, and still struggles with extremities such as fingers and ankles due to marker sparsity~\cite{DBLP:journals/corr/abs-2012-06178,Chen20223DHB}. These limitations collectively hinder the development of robust, high-precision whole-body reconstruction models, most notably in challenging scenarios such as depth ambiguity, self-occlusion, and highly articulated body movements.

To address these limitations, we introduce Human4K, a pioneering real-world whole-body motion capture dataset recorded entirely in native 4K resolution and composed of 6 million multi-view images. Human4K is captured in a controlled indoor studio using a synchronized eight-camera 4K acquisition system paired with a professional marker-based Vicon mocap setup\footnote{https://www.vicon.com/.}, ensuring precise spatial and temporal alignment between imagery and ground-truth 3D motion. The dataset features 11 professional actors and dancers performing a broad range of actions that span daily behaviors, complex full-body movements, and highly challenging scenes involving self-occlusion, depth ambiguity, and extreme poses. All motion sequences are processed through a \textbf{M}otion-\textbf{R}etargeting and \textbf{R}efinement \textbf{M}odule (MRRM) that produces high-accuracy SMPL-X annotations, with particular attention to extremities such as hands and ankles. Through this design, Human4K provides a uniquely comprehensive combination of 4K visual detail, mocap-level annotation fidelity, and rich whole-body motion diversity, offering strong support for training high-precision 3D human reconstruction models. Extensive experiments further demonstrate that models trained on Human4K achieve clear and consistent improvements across multiple benchmarks, confirming its effectiveness and practical value.

Our contribution can be summarized as follows:
\begin{itemize}
\item We present Human4K, a large-scale whole-body reconstruction dataset captured in native 4K, featuring 6 million synchronized multi-view images with mocap-accurate SMPL-X annotations and diverse complex and extreme motions.

\item We develop MRRM, a motion-retargeting and refinement module that accurately converts Vicon mocap data into SMPL-X parameters, enabling high-precision alignment of the whole body and extremities.

\item Extensive experiments show that Human4K markedly improves reconstruction accuracy, especially in challenging regions involving extremities and depth ambiguity.
\end{itemize}

\section{Related Works}
\subsection{Human Reconstruction Datasets}
3D whole-body human reconstruction is fundamental for building high-fidelity digital humans, supporting applications such as audio-driven facial animation and multimodal avatars~\cite{han2024pmmtalk, han2025pestalk}. Existing datasets can be broadly grouped into four categories, covering pseudo-3D annotations, synthetic renderings, multiview captures and marker-based mocap. While they have collectively driven progress, they still lack the geometric precision and motion diversity needed for robust whole-body modeling, especially for extremities and complex, depth-ambiguous poses.

As summarized in Table~\ref{table1}, pseudo-3D datasets such as MSCOCO~\cite{lin2014microsoft}, MPII~\cite{andriluka20142d}, LSPET~\cite{johnson2010clustered}, OCHuman~\cite{zhang2019pose2seg}, Ubody~\cite{lin2023one}, and H3WB~\cite{zhu2023h3wb} lift 2D keypoints to SMPL/SMPL-X parameters via annotators like NeuralAnnot~\cite{Moon_2022_CVPRW_NeuralAnnot}, which makes this strategy easy to scale but noisy and coarse, particularly for ankles, wrists and fingers. Synthetic datasets such as AGORA~\cite{Patel:CVPR:2021}, SynBody~\cite{yang2023synbody}, BEDLAM~\cite{black2023bedlam}, GTA-Human~\cite{cai2021playing} and SURREAL~\cite{varol17_surreal} provide clean labels and full control, but suffer from a large domain gap to real images. Multiview datasets such as HuMMan~\cite{cai2022humman}, MVhumanNet~\cite{xiong2024mvhumannet}, Genebody~\cite{cheng2022generalizable}, MPI-INF-3DHP~\cite{mono-3dhp2017} and ZJUmocap~\cite{easymocap}, together with multi-camera reconstruction and tracking systems that focus on outdoor or real-time capture~\cite{zhu2016video, liu2016human, alexiadis2016integrated}, collect large-scale real data with many cameras. However, their annotations still rely on multiview optimization or pseudo-labeling and remain unreliable under severe occlusion or extreme articulation. Mocap datasets including Human3.6M~\cite{6682899}, HumanEVA~\cite{sigal2006humaneva}, 3DPW~\cite{vonMarcard2018}, AMASS~\cite{AMASS:ICCV:2019}, EHF~\cite{SMPL-X:2019}, EMDB~\cite{kaufmann2023emdb} and RICH~\cite{Huang:CVPR:2022} offer accurate 3D ground truth via markers, but typically lack high-resolution imagery, have limited motion types or provide incomplete extremity coverage.

Given these limitations, there is still no real-world dataset that jointly offers native 4K imagery, mocap-accurate full-body SMPL-X supervision, and diverse whole-body motions with extreme articulation, depth ambiguity, and strong self-occlusion. Our Human4K dataset is designed to fill this gap by providing a large-scale 4K multi-view mocap corpus with high-precision annotations and rich whole-body movements, enabling more reliable and fine-grained reconstruction of complex human poses.

\subsection{Parametric Human Body Models}
Parametric human body models provide a compact and differentiable representation of human shape and pose, making them widely used in vision and graphics applications. SMPL~\cite{SMPL:2015} is a learned body model with shape bases and pose-dependent blendshapes, offering an effective balance between realism and computational efficiency. To better capture fine-grained details, MANO~\cite{MANO:SIGGRAPHASIA:2017} and FLAME~\cite{FLAME:SiggraphAsia2017} were later proposed to model hand articulation and facial expressions, respectively. Building on these components, SMPL-X~\cite{SMPL-X:2019} integrates the body, hands and face into a unified full-body model with articulated extremities and expressive motion. Owing to its expressiveness and compatibility, SMPL-X is now a standard choice for whole-body reconstruction and is therefore adopted in Human4K.

\subsection{3D Human Reconstruction Methods}
Recent 3D human reconstruction methods focus on improving robustness under challenging imaging conditions. Zou et al.~\cite{zou2022human} estimate human pose and shape from single polarization images, while Wang et al.~\cite{wang2023global} and Li et al.~\cite{li2022exploiting} design global–local and temporal transformer architectures for video-based 3D pose estimation. In parallel, anatomy-aware and spatio–temporal modeling have also been extensively explored. Chen et al.~\cite{chen2021anatomy} incorporate bone-based pose decomposition to better respect human skeletal structure, Tang et al.~\cite{tang2023ftcm} propose a frequency–temporal collaborative module for video 3D pose estimation, Yao et al.~\cite{yao2024staf} fuse spatial and temporal cues for 3D human mesh recovery from monocular video, and Hu et al.~\cite{hu2023personalized} introduce personalized graph generation to capture subject-specific pose and shape characteristics. For high-fidelity geometry, Liu et al.~\cite{liu2024seif} and Wang et al.~\cite{wang2021prior} reconstruct detailed 3D heads from monocular or multi-view RGB images with strong priors.

Building on these advances, we focus on SMPL-X-based whole-body reconstruction. In this work, we consider three representative methods: the multi-stage Hand4Whole~\cite{Moon_2022_CVPRW_Hand4Whole} and the one-stage OSX-b~\cite{lin2023one} and SMPLer-X-b~\cite{cai2023smplerxscalingexpressivehuman}. All these approaches benefit from accurate geometric supervision and diverse poses. Human4K further supports them by supplying native 4K multi-view images with mocap-accurate SMPL-X annotations for the full body, hands, and face.

\section{Human4K Dataset}
\textbf{Overview.} Human4K is a large-scale whole-body motion capture dataset designed to support high-precision 3D human reconstruction research. As illustrated by Fig.~\ref{fig3}, raw motion is captured by an eight-camera 4K system together with a 120 FPS Vicon setup. The two data streams are temporally aligned through high-precision timestamps, after which the recorded motion is exported in BVH format. The BVH sequences are then retargeted to the SMPL-X model to obtain coherent whole-body pose sequences. A dedicated hand-optimization stage is applied to correct structural mismatches between mocap skeletal data and the SMPL-X hand model, while facial expressions are obtained through a separate estimation process, enabling complete full-body annotations that cover body, hands, and face.

Finally, the Human4K dataset contains 6,007,290 synchronized 4K frames from 11 professional actors performing diverse daily activities together with challenging motions that exhibit strong self-occlusion and depth ambiguity. Since all recordings are conducted in motion-capture suits to ensure reliable marker tracking, we introduce an additional virtual clothing augmentation process applied to half of the dataset. This augmentation enriches appearance variation while preserving identity, geometry, and pose consistency, which enhances the dataset’s applicability to real-world scenarios.

The following sections provide a detailed description of each stage of the Human4K construction pipeline. We begin with the multi-view capture system and synchronization mechanisms before describing the motion-retargeting and refinement modules, the virtual clothing augmentation strategy, and the statistical characteristics of the dataset.

\subsection{Motion Data Acquisition}
\subsubsection{Multi-view Camera Setup}
\begin{figure}[!t]
\centering
\includegraphics[width=\linewidth]{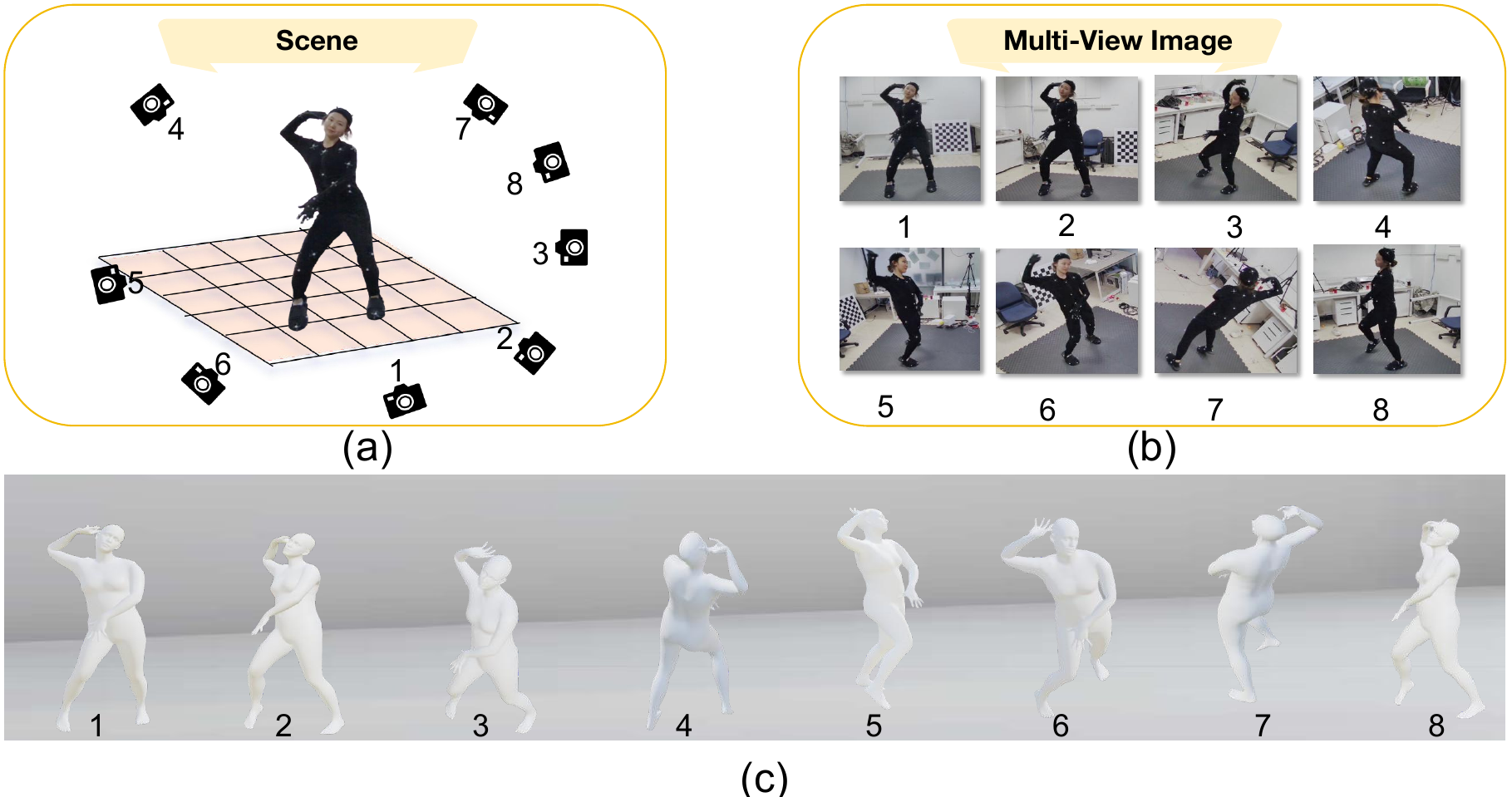}
\caption{Overview of the Human4K capture setup. (a) Eight 4K cameras placed around the subject. (b) Synchronized multi-view RGB images. (c) Reconstructed 3D mesh, revealing complex articulation and self-occlusion (e.g., views 5 and 8).}
\label{fig1}
\end{figure}

To capture full-body human motion with strong geometric constraints, we employ an eight-camera 4K configuration arranged around the capture volume, as shown in Fig.~\ref{fig1}. The cameras are positioned at the front, back, left, right, and four diagonal viewpoints, providing dense angular coverage for challenging poses involving self-occlusion and large limb articulation. This setup forms the foundation of our multi-view RGB data and enables reliable supervision for SMPL-X annotation.

\subsubsection{Image Capturing System}
Building on this camera layout, we develop a distributed acquisition system capable of recording synchronized 4K streams from all eight viewpoints. The system consists of a central controller that broadcasts periodic timing signals and multiple client machines responsible for camera control and image encoding. Each client handles up to three cameras and performs signal reception, triggering, and GPUJPEG-based 4K compression in parallel threads. This distributed architecture eliminates bandwidth bottlenecks on a single machine and supports stable multi-view 4K recording at up to 15\,FPS without frame drops.

\subsubsection{Timestamp Alignment}
The Vicon mocap system records motion at 120\,FPS, whereas the cameras operate at 15\,FPS. To merge these asynchronous sources into a unified dataset, each RGB frame and mocap sample is stored with a precise timestamp. During preprocessing, every image is paired with the temporally closest Vicon frame, producing a reliable one-to-one correspondence between visual observations and 3D motion states. This alignment is essential for training and evaluating models that require frame-level synchronized supervision.

\begin{figure}[!t]  
\centering
\includegraphics[width=\linewidth]{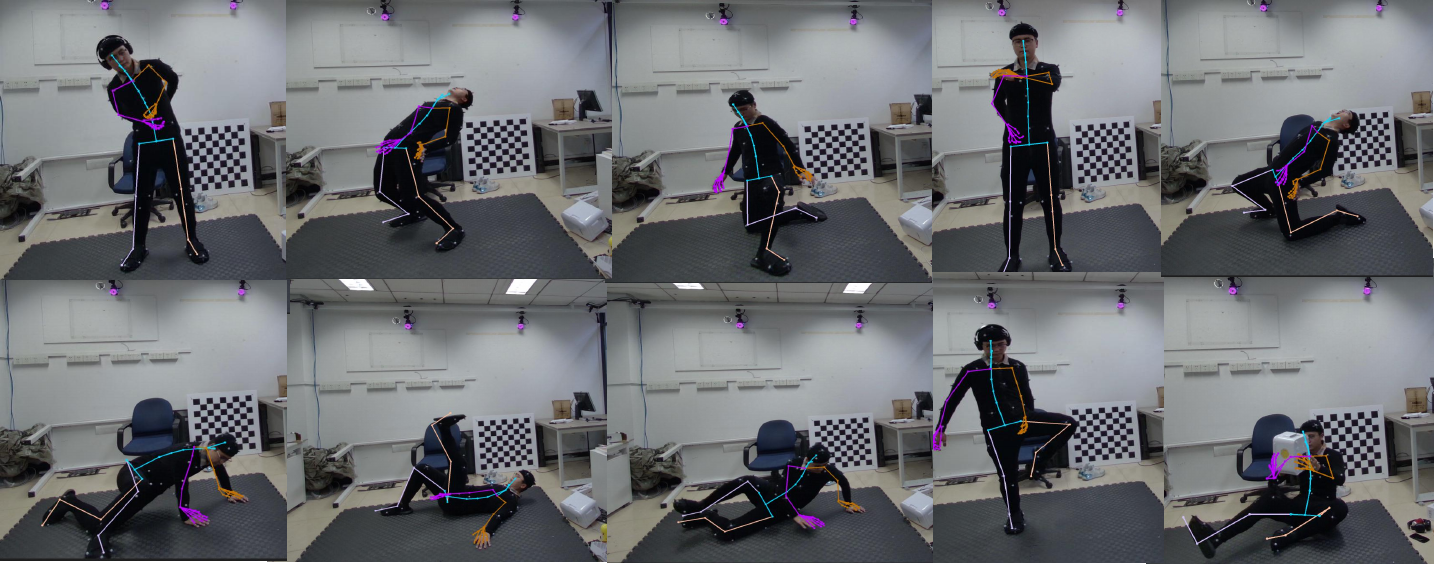}
\caption{Projected 3D Vicon joints overlaid on 2D image observations, illustrating the accurate alignment achieved by our camera calibration.}
\label{fig2}
\end{figure}

\begin{figure*}[t]
\centering
\includegraphics[width=0.9\linewidth, height=8.5cm]{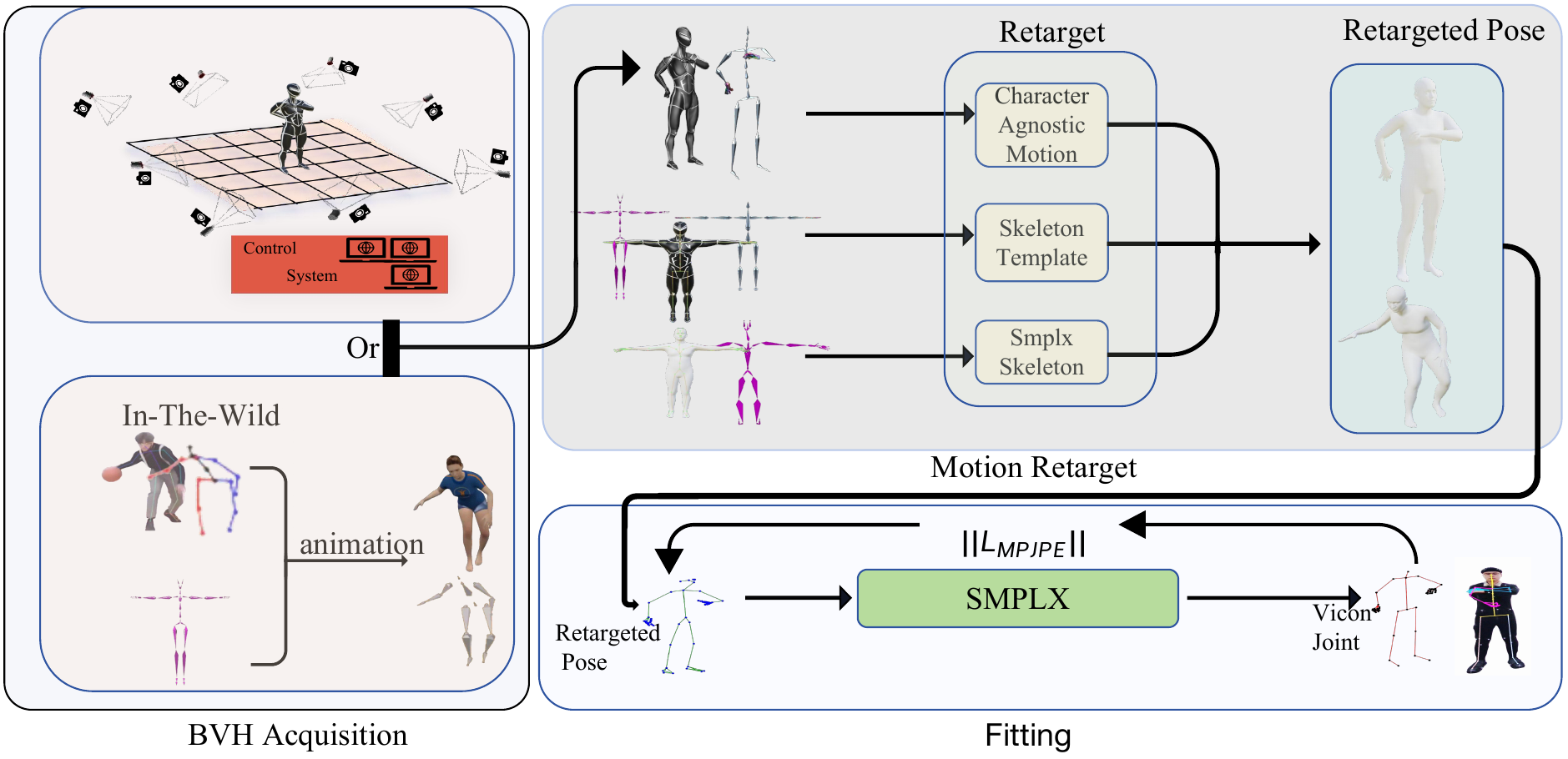}
\caption{\textbf{Pipeline for motion data acquisition and SMPL-X pose retargeting.} Motion data can be obtained from a Vicon mocap system (used in Human4K) or from in-the-wild video via 3D pose estimation, converted to BVH format, retargeted to the SMPL-X skeleton, and further refined for accurate full-body and hand annotations.}
\label{fig3}
\end{figure*}

\subsection{Motion-Retargeting and Refinement Module}
To convert raw mocap data into high-quality SMPL-X annotations, we design a Motion-Retargeting and Refinement Module (MRRM) with four stages: (1) camera calibration via 2D–3D joint alignment, (2) skeleton retargeting to the SMPL-X kinematic tree, (3) hand-specific optimization to correct structural mismatches, and (4) facial parameter extraction and integration. This pipeline ensures that the final SMPL-X parameters tightly match the Vicon mocap trajectories for both body and extremities.

\subsubsection{Camera Calibration via 2D–3D Joint Alignment}
To ensure geometric consistency between mocap data and the recorded images, we first recover the camera parameters for each viewpoint. Accurate extrinsics are crucial, as all subsequent annotations must align with the 4K multi-view imagery, otherwise the projected 3D joints would deviate from their image observations.

We estimate camera poses by projecting 3D Vicon joints onto each view. A 2D pose estimator provides image-space joints, which are paired with corresponding 3D joints and fed into OpenCV’s \texttt{solvePnP} to solve for the camera rotation $\mathbf{R}_{cam}$ and translation $\mathbf{T}_{cam}$. With the recovered extrinsics $(\mathbf{R}_{cam}, \mathbf{T}_{cam})$ and intrinsics $\mathbf{K}$, each 3D keypoint $\mathbf{P}_w$ is projected using the standard pinhole model:
\[
\mathbf{P}_c = \mathbf{R}_{cam}\mathbf{P}_w + \mathbf{T}_{cam}, \quad
\lambda \begin{bmatrix} x \\ y \\ 1 \end{bmatrix} = \mathbf{K}\mathbf{P}_c,
\]
where $\mathbf{P}_c$ is the point in the camera coordinate system, $(x,y)$ is its 2D image coordinate, and $\lambda$ is a scaling factor in homogeneous coordinates. The projection results in Fig.~\ref{fig2} show that the 3D Vicon joints align well with the 2D detections, which confirms the accuracy of the recovered camera parameters.

\subsubsection{Motion Retargeting and Optimization}
Although Vicon mocap provides precise 3D joint trajectories in BVH format, its skeletal structure and bone proportions differ from those of SMPL-X. To obtain coherent SMPL-X pose parameters from these BVH sequences, we adopt a motion retargeting pipeline that systematically resolves these structural discrepancies and produces accurate full-body pose annotations.

Our pipeline consists of four steps. 
\textbf{(1) Skeleton mapping.} We first establish a joint-level correspondence between the Vicon skeleton and SMPL-X using a standardized T-pose to ensure anatomical consistency across the head, spine, and limbs. 
\textbf{(2) Bone proportion adjustment.} Because the two skeletons differ in bone lengths, the source joints are scaled and aligned to the target skeleton according to
\[
\mathbf{p}_i^T = s\,\mathbf{M}\,\mathbf{p}_i^S,
\]
where $\mathbf{p}_i^S$ and $\mathbf{p}_i^T$ denote source and target joint positions, $s$ is a global scale factor, and $\mathbf{M}$ aligns bone directions.
\textbf{(3) Pose correction and rotation transfer.} Joint rotations are transformed to maintain plausible articulation in SMPL-X:
\[
\mathbf{R}_i^T = \mathbf{M}\,\mathbf{R}_i^S\,\mathbf{M}^{-1},
\]
ensuring consistent orientation across skeletons.
\textbf{(4) Pose parameter extraction.} Finally, canonical-space SMPL-X rotations are computed as
\[
\widehat{\mathbf{R}}_i(t) = 
\widehat{\mathbf{R}}_i(0)\,
\widehat{\mathbf{R}}_{i,\text{src}}(t)\,
\widehat{\mathbf{R}}^{-1}_{i,\text{src}}(0),
\]
where $\widehat{\mathbf{R}}_i(0)$ is the T-pose rotation.

It is worth noting that the retargeting pipeline is not limited to BVH sequences obtained from indoor Vicon captures. As illustrated in the BVH Acquisition block of Fig.~\ref{fig3}, our framework can also operate on motion data extracted from in-the-wild videos. In such cases, 3D pose estimators are applied to monocular footage to obtain joint trajectories, which are then converted into BVH-style skeletal motion before being processed by our retargeting module. This flexibility allows the pipeline to handle both controlled mocap recordings and unconstrained real-world motion, broadening its applicability across diverse environments.

\subsubsection{SMPL-X Annotation} The retargeted skeleton provides the basis for constructing full SMPL-X parameters. We extract the global body rotation $\theta_{g}^{b} \in \mathbb{R}^{3}$, body joint rotations $\theta_{b} \in \mathbb{R}^{21 \times 3}$, and initial hand rotations $\theta_{h} \in \mathbb{R}^{15 \times 3}$. The shape parameters $\beta_{b} \in \mathbb{R}^{10}$ are initialized to an average template, and facial parameters are temporarily set to zero to focus on body dynamics. Because the Vicon hand skeleton differs substantially from the SMPL-X hand structure, naive retargeting leads to suboptimal hand articulation, motivating an additional hand-specific optimization stage.

\begin{table}[t]
\centering
\scriptsize
\setlength{\tabcolsep}{3pt}
\renewcommand{\arraystretch}{1.2}
\caption{\textbf{Comparison of retargeting and optimized fitting.} All metrics are in millimeters (mm).}
\begin{tabular}{lcccccc}
\toprule
\multirow{2}{*}{\textbf{Method}} &
\multicolumn{2}{c}{\textbf{Whole}} &
\multicolumn{2}{c}{\textbf{Body}} &
\multicolumn{2}{c}{\textbf{Hand}} \\
\cmidrule(r){2-3}\cmidrule(lr){4-5}\cmidrule(l){6-7}
& MPJPE & PA-MPJPE & MPJPE & PA-MPJPE & MPJPE & PA-MPJPE \\
\midrule
Retarget & 27.8 & 25.5 & 32.3 & 31.8 & 25.5 & 20.7 \\
Retarget + Opt. & \textbf{16.8} & \textbf{14.2} & \textbf{27.5} & \textbf{25.2} & \textbf{12.5} & \textbf{8.4} \\
\bottomrule
\end{tabular}
\label{tab2}
\end{table}

\begin{figure}[t]
\centering
\includegraphics[width=0.9\linewidth]{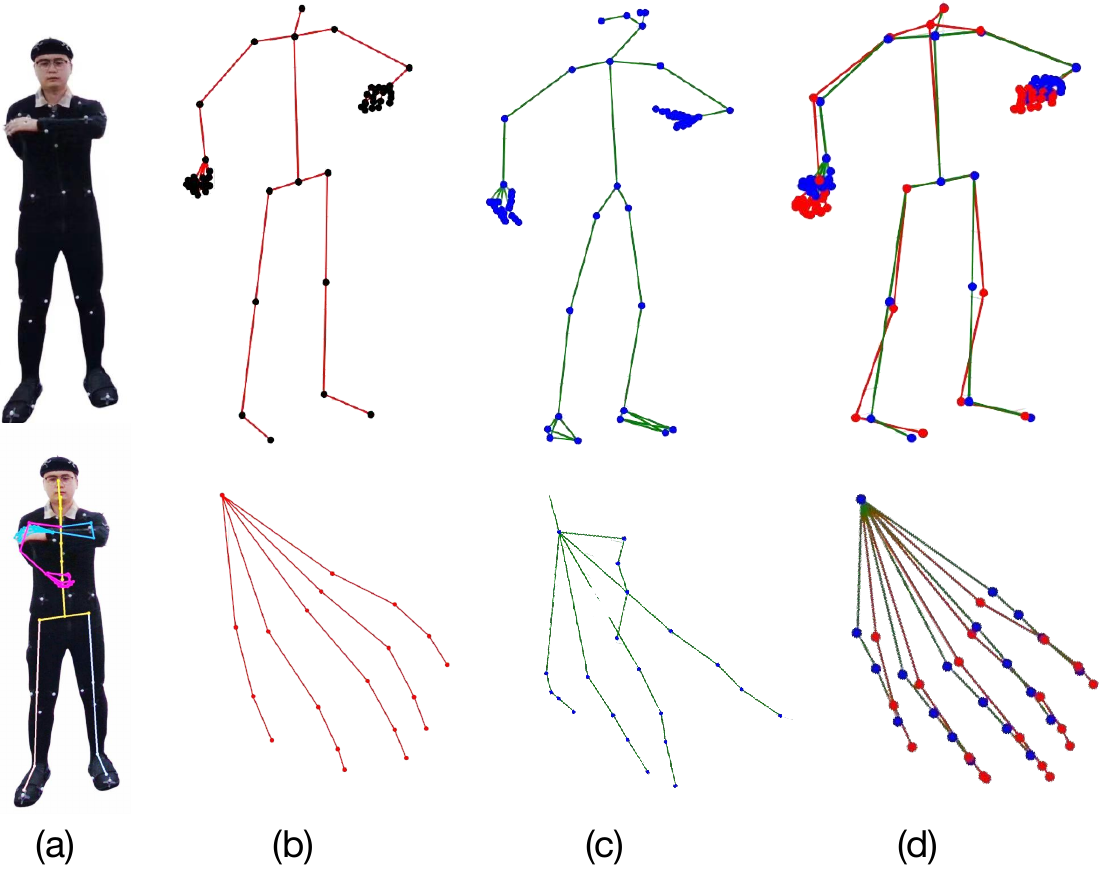}
\caption{Comparison of hand pose quality before and after fitting. 
(a) Human action input; (b) Vicon mocap joints; 
(c) SMPL-X pose after motion retargeting; 
(d) refined SMPL-X pose after hand-fitting, showing improved alignment and articulation.}
\label{fig4}
\end{figure}

\textbf{Hand Pose Fitting.}
Upon examining the retargeted skeleton, we observed noticeable inaccuracies in the reconstructed hand poses due to the structural mismatch between the Vicon hand skeleton and the SMPL-X hand model. To resolve these discrepancies, we perform a dedicated hand-fitting stage in which the left- and right-hand pose parameters, $\theta_{lh}, \theta_{rh} \in \mathbb{R}^{15 \times 3}$, are treated as learnable variables.

During optimization, the SMPL-X model generates 3D hand joint positions from the current pose estimates, which are then supervised using the corresponding Vicon joint locations. We employ the Mean Per Joint Position Error (MPJPE) as the objective function, minimizing the Euclidean distance between predicted and ground-truth joints to enforce accurate geometric alignment. The parameters are optimized using the Adam optimizer with an initial learning rate of $1\times 10^{-2}$ for 3000 iterations. Adam’s adaptive updates facilitate both rapid early-stage convergence and fine-grained refinement in the later stages. We additionally allow minor adjustments to shape parameters to better accommodate subject-specific hand proportions.

As shown in Table~\ref{tab2}, the hand-fitting stage substantially improves the accuracy of aligning the SMPL-X hand joints with the Vicon ground-truth markers. Since this evaluation is performed directly on the Human4K mocap data, the reported MPJPE and PA-MPJPE reflect the quality of the retargeting procedure rather than the performance of a learned reconstruction model. After optimization, the hand MPJPE is reduced to 12.5\,mm and the PA-MPJPE to 8.4\,mm, indicating a much closer correspondence between the reconstructed and captured hand motions. Figure~\ref{fig4} further illustrates this improvement, showing that the optimized hand poses better follow the underlying Vicon trajectories and exhibit more anatomically plausible articulation.

\textbf{Facial Parameter Extraction.} For facial data, we first use YOLOv8 to detect and crop face regions from the 4K images. The cropped faces are then processed by DECA~\cite{feng2021learning}, which predicts FLAME parameters, including expression coefficients $\psi \in \mathbb{R}^{10}$ and jaw pose $\theta_f \in \mathbb{R}^3$. These parameters are directly compatible with the SMPL-X model, enabling seamless integration of facial expressions into the full-body representation.

After obtaining the pseudo-ground-truth parameters for the body, hands, and face, we generate the final whole-body 3D pseudo-GTs by feeding
\[
\{ \theta_g^b,\, \theta_b,\, \beta_b,\, \theta_{rh},\, \theta_{lh},\, \theta_f,\, \psi \}
\]
into the SMPL-X model. This unified parameter set ensures consistent geometric structure across the entire body and supports high-quality full-body annotation.

\begin{figure}[!t]
\centering
\includegraphics[width=\linewidth]{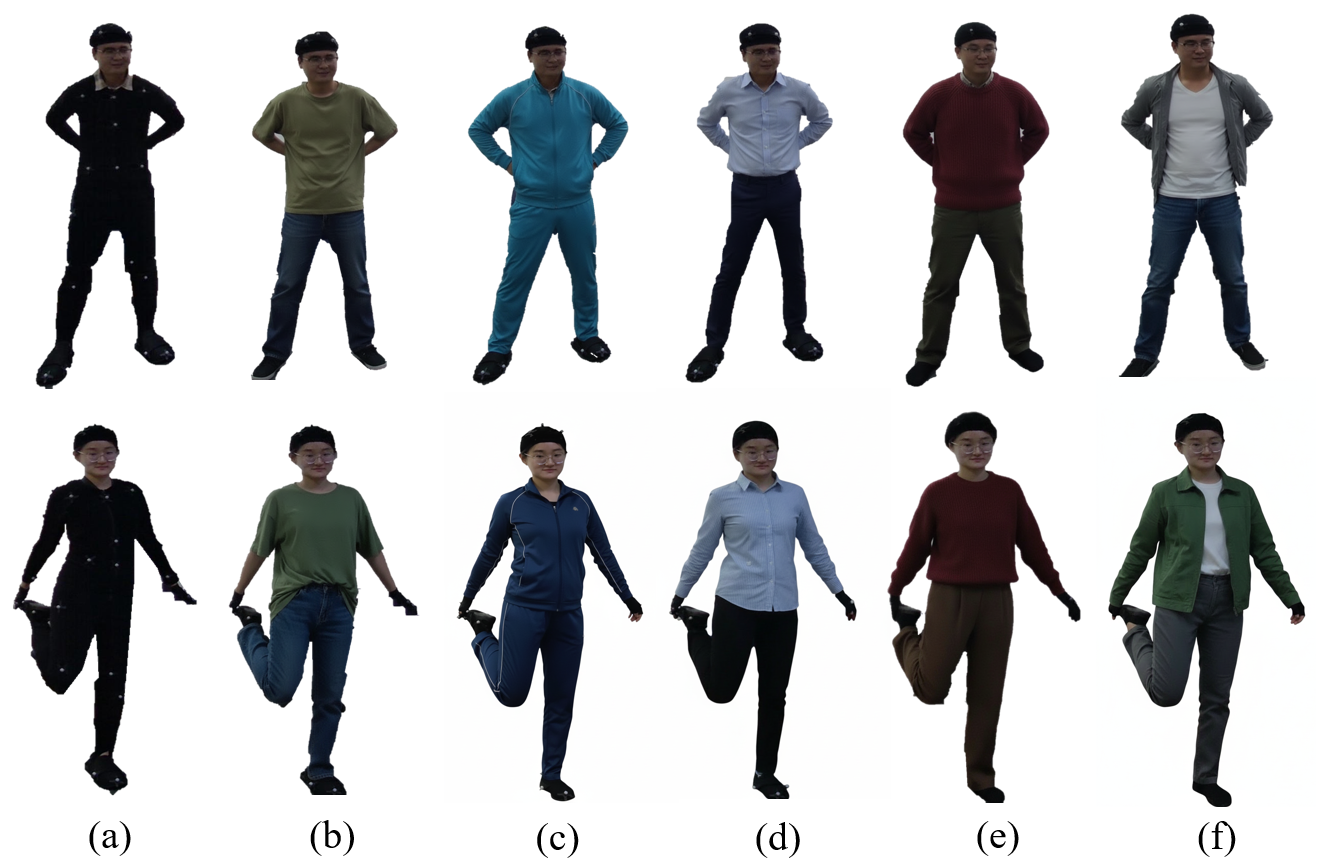}
\caption{
Examples of virtual clothing augmentation applied to Human4K.  
Each row contains six images, where (a)–(f) show the original RGB frames and the remaining columns present the corresponding augmented results for the following outfit categories:  
(1) loose T-shirt with jeans ,  
(2) athletic tracksuit,  
(3) button-up shirt with slim-fit pants,  
(4) crewneck sweater with casual pants, and  
(5) casual jacket with long pants.  
These clothing variants introduce diverse textures and materials while preserving identity, pose, and body shape.
}
\label{fig5}
\end{figure}

\subsection{Synthetic Clothing Augmentation for Appearance Diversity}
Although Human4K provides high-precision mocap supervision and extensive whole-body motion, all recordings are performed in standardized mocap suits. This design ensures reliable marker tracking but limits appearance diversity, especially in clothing style, texture, and color, which may constrain model generalization under real-world dressing conditions.

To alleviate this issue, we apply virtual clothing augmentation to 50\% of the data using Google's Nano model\footnote{https://aistudio.google.com/}. The goal is to enrich clothing appearance while preserving human identity, body shape, and pose. We design five representative everyday clothing categories that cover common variations in garment fit, material, and seasonality:

\begin{itemize}
    \item \textbf{Loose T-shirt + jeans (casual).}  
    T-shirts in random solid colors are paired with blue or dark-washed jeans, adding variation in tone and fabric texture.

    \item \textbf{Athletic tracksuit (sportswear).}  
    Jacket and pants in realistic sports colors introduce stretch fabrics and subtle sheen effects.

    \item \textbf{Button-up shirt + slim-fit pants (commuting).}  
    Solid or lightly patterned shirts with neutral-tone trousers provide more structured apparel geometry.

    \item \textbf{Crewneck sweater + casual pants (fall/winter).}  
    Knit sweaters in warm earth tones increase seasonal diversity and introduce distinct texture patterns.

    \item \textbf{Casual jacket + long pants (outdoor).}  
    Lightweight jackets create layered upper-body silhouettes and richer garment folds.
\end{itemize}

Across these categories, garment colors and textures are randomized, while identity, pose, and body proportions remain unchanged. This augmentation substantially increases appearance diversity in Human4K and improves the robustness of models trained on the dataset to real-world clothing variation, without compromising the geometric consistency required for accurate SMPL-X supervision. Representative examples are shown in Fig.~\ref{fig5}.

\begin{table}[t]
\centering
\setlength{\tabcolsep}{16pt}
\renewcommand{\arraystretch}{1.2}
\caption{\textbf{Frame statistics of Human4K.}
The table lists training, validation, and test frame counts for each scenario, highlighting the dataset’s scale and motion diversity.}
\begin{tabular}{@{}lccc@{}}
\toprule
\textbf{Scenario} & \textbf{Train} & \textbf{Val} & \textbf{Test} \\
\midrule

\rowcolor{gray!10}
\multicolumn{4}{c}{\textbf{Upbody}} \\
\quad discuss & 370{,}096 & 96{,}520 & 95{,}168 \\
\quad phone   & 371{,}008 & 94{,}096 & 95{,}680 \\

\rowcolor{gray!10}
\multicolumn{4}{c}{\textbf{Sitting on Chair}} \\
\quad eating  & 382{,}360 & 93{,}312 & 98{,}808 \\
\quad reading & 366{,}024 & 93{,}488 & 94{,}904 \\

\rowcolor{gray!10}
\multicolumn{4}{c}{\textbf{Sitting on Floor}} \\
\quad posing   & 367{,}392 & 90{,}648 & 91{,}408 \\
\quad clothing & 374{,}368 & 91{,}976 & 97{,}427 \\

\rowcolor{gray!10}
\multicolumn{4}{c}{\textbf{Lying}} \\
\quad relax   & 335{,}672 & 91{,}472 & 89{,}456 \\
\quad phoning & 321{,}256 & 95{,}712 & 97{,}544 \\

\rowcolor{gray!10}
\multicolumn{4}{c}{\textbf{Moving}} \\
\quad walking & 328{,}136 & 92{,}928 & 97{,}664 \\

\rowcolor{gray!10}
\multicolumn{4}{c}{\textbf{Sporting}} \\
\quad sporting & 327{,}376 & 93{,}392 & 96{,}105 \\

\rowcolor{gray!10}
\multicolumn{4}{c}{\textbf{Dance}} \\
\quad dancing  & 390{,}654 & 91{,}952 & 93{,}288 \\
\midrule

\textbf{Total} & 3{,}934{,}342 & 1{,}025{,}496 & 1{,}047{,}452 \\
\bottomrule
\end{tabular}
\label{tab3}
\end{table}

\begin{figure}[!t]
\centering
\includegraphics[width=\linewidth]{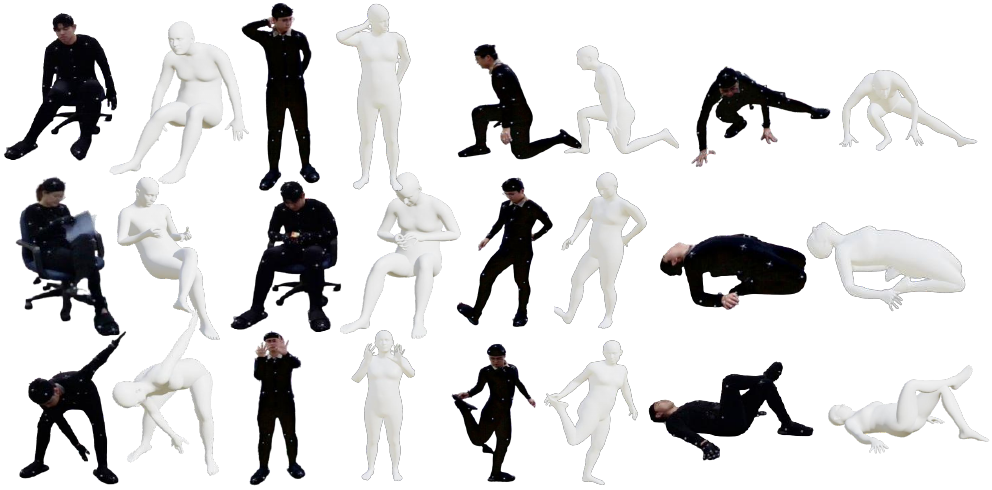}
\caption{Sample frames from the Human4K dataset, illustrating diverse motions and accurate full-body annotations.}
\label{fig6}
\end{figure}

\subsection{Statistical Information}
Human4K contains recordings of 11 professional actors and dancers (7 males and 4 females) with heights ranging from 157 to 180 cm and BMIs between 17 and 27, covering a broad spectrum of body shapes. The subjects are divided into training/validation (6 males and 3 females) and test sets (1 male and 1 female). In total, the dataset provides 6,007,290 synchronized 4K frames captured by eight cameras across 11 improvised scenarios, including standing, sitting, exercising, eating, and object interaction. Detailed frame counts for each scenario are summarized in Table~\ref{tab3}.

As illustrated in Fig.~\ref{fig2}, the multi-view 4K images are tightly aligned with the reconstructed 3D poses, demonstrating the high geometric fidelity achieved by our capture setup. The dataset features native 4K resolution and a rich variety of complex full-body motions, including self-occlusion and severe depth ambiguity, which are rarely covered in existing resources. These characteristics make Human4K a highly challenging and comprehensive benchmark suitable for a wide range of applications, such as whole-body pose estimation, motion analysis, virtual reality, and character animation.

\begin{table*}[!t]
\centering
\setlength{\tabcolsep}{12pt}
\renewcommand{\arraystretch}{1.35}
\caption{Comparison of 3D body reconstruction errors with and without Human4K augmentation. We report metrics for three methods (Hand4Whole~\cite{Moon_2022_CVPRW_Hand4Whole}, OSX-b~\cite{lin2023one}, SMPLer-X-b) when trained on public datasets (P: COCO-WholeBody~\cite{lin2014microsoft}, MPII~\cite{andriluka20142d}, Human3.6M~\cite{6682899}, MPI-INF-3DHP~\cite{mono-3dhp2017}), on Human4K alone (Ours), and on the combined set (P+Ours).}
\begin{tabular}{lccccccccc}
\toprule
\multirow{2}{*}{\textbf{Method}} & \multirow{2}{*}{\textbf{Train set}} &
\multicolumn{3}{c}{\textbf{EHF MPVPE (mm)}$\downarrow$} &
\multicolumn{3}{c}{\textbf{EHF PA-MPVPE (mm)}$\downarrow$} &
\multicolumn{2}{c}{\textbf{3DPW Body (mm)}$\downarrow$} \\
\cmidrule(lr){3-5} \cmidrule(lr){6-8} \cmidrule(lr){9-10}
& & All & Hand & Face & All & Hand & Face & MPJPE & PA-MPJPE \\
\midrule

\multirow{3}{*}{Hand4Whole~\cite{Moon_2022_CVPRW_Hand4Whole}}
 & P        & 79.00 & 44.11 & 23.99 & 51.29 & 13.13 & 5.79 & 102.12 & 68.48 \\
 & Ours     & 82.16 & 41.29 & 23.51 & 53.34 & 12.34 & 5.67 & 105.18 & 70.53 \\
 & P+Ours   & \textbf{70.10} & \textbf{40.12} & \textbf{22.13} & \textbf{47.26} & \textbf{10.77} & \textbf{5.24} & \textbf{95.02} & \textbf{63.05} \\
\midrule

\multirow{3}{*}{OSX-b~\cite{lin2023one}}
 & P        & 81.29 & 60.43 & 26.58 & 52.85 & 17.79 & 5.98 & 98.58 & 64.93 \\
 & Ours     & 84.54 & 56.00 & 26.05 & 55.13 & 16.72 & 5.86 & 101.54 & 66.87 \\
 & P+Ours   & \textbf{72.31} & \textbf{48.98} & \textbf{23.62} & \textbf{46.03} & \textbf{14.81} & \textbf{5.33} & \textbf{90.29} & \textbf{59.12} \\
\midrule

\multirow{3}{*}{SMPLer-X-b~\cite{cai2023smplerxscalingexpressivehuman}}
 & P        & 82.12 & 79.98 & 32.30 & 58.14 & 40.11 & 15.95 & 88.30 & 60.54 \\
 & Ours     & 85.40 & 75.18 & 31.66 & 60.47 & 37.70 & 15.64 & 91.83 & 62.36 \\
 & P+Ours   & \textbf{75.92} & \textbf{71.24} & \textbf{26.94} & \textbf{50.93} & \textbf{37.13} & \textbf{12.11} & \textbf{81.21} & \textbf{56.01} \\
\bottomrule
\end{tabular}
\label{tab4}
\end{table*}

\section{Experiments}
In this section, we present a systematic evaluation of Human4K. \textbf{(1) Enhancing model performance across datasets.} We first investigate how Human4K improves existing SMPL-X based reconstruction methods on standard benchmarks and examine whether its high-fidelity annotations provide complementary supervision that benefits cross-dataset generalization. \textbf{(2) Establishing Human4K as a challenging in-domain benchmark.} We then evaluate Human4K as an in-domain test set and analyze the challenges introduced by its 4K resolution, extensive whole-body motion range, strong self-occlusions, and depth ambiguity. \textbf{(3) Characterizing motion and pose diversity.} We compare the pose distribution of Human4K with those of prior datasets in order to understand the uniqueness, coverage, and diversity of its full-body motion patterns. \textbf{(4) Understanding the impact of 4K resolution.} We further study how high-resolution imagery affects reconstruction accuracy and analyze the quantitative advantages provided by 4K visual inputs. Finally, we provide qualitative visualizations to highlight the geometric fidelity, annotation accuracy, and visual complexity of Human4K.

\subsection{Experimental Setup}
\noindent\textbf{Evaluation Metrics.}
We adopt four standard metrics to quantitatively evaluate 3D human reconstruction quality. 
MPJPE (Mean Per Joint Position Error) and PA-MPJPE (Procrustes Aligned MPJPE) measure the accuracy of predicted 3D joints before and after rigid alignment. 
Similarly, MPVPE (Mean Per Vertex Position Error) and PA-MPVPE (Procrustes Aligned MPVPE) assess the consistency of reconstructed meshes at the vertex level. 
These metrics jointly reflect both global body alignment and local geometric fidelity. 
All results are reported in millimeters to enable precise and fair comparison across different methods and datasets.

\noindent\textbf{Benchmark Datasets.}
For cross-dataset generalization, we evaluate on two widely used benchmarks that emphasize complementary aspects of whole-body reconstruction. 
3DPW~\cite{vonMarcard2018} is used to assess body and whole-body performance in challenging in-the-wild scenes with natural backgrounds and unconstrained motion. 
EHF~\cite{SMPL-X:2019} focuses on detailed full-body evaluation and provides high-quality annotations for expressive hands and faces, making it suitable for analyzing extremity and facial accuracy. 
In addition, Human4K is itself partitioned into training, validation, and test subsets as described in the dataset section. 
We use its test split as an in-domain benchmark to measure performance under the same 4K multi-view and mocap conditions that the dataset is designed to capture.

\noindent\textbf{Baselines and Training Sets.}
We consider three representative SMPL-X based methods that cover both multi-stage and one-stage designs. 
Hand4Whole~\cite{Moon_2022_CVPRW_Hand4Whole} adopts a multi-stage pipeline with dedicated components for hand modeling and is further trained with additional hand supervision from InterHand2.6M~\cite{Moon_2020_ECCV_InterHand2.6M} and FreiHAND~\cite{Freihand2019}. 
In contrast, OSX-b~\cite{lin2023one} and SMPLer-X-b~\cite{cai2023smplerxscalingexpressivehuman} use one-stage architectures that directly predict whole-body SMPL-X parameters from images.

Following prior work, we use COCO-WholeBody~\cite{lin2014microsoft}, MPII~\cite{andriluka20142d}, Human3.6M~\cite{6682899}, and MPI-INF-3DHP~\cite{mono-3dhp2017} as the standard training combination, denoted as \textbf{P}. In our notation, \textbf{Ours} refers to models trained exclusively on Human4K, while \textbf{P+Ours} indicates joint training on the public datasets and Human4K. This configuration allows us to isolate the effect of Human4K as a standalone data source and to quantify its benefit when used as an additional high-fidelity supervision signal on top of existing benchmarks.

\subsection{Enhancing Model Performance Across Datasets}
A central question of our experimental analysis is whether Human4K can improve the performance of existing SMPL-X based methods on standard benchmarks. Therefore, we evaluate cross-dataset performance on EHF and 3DPW, and Table~\ref{tab4} reports the reconstruction errors of Hand4Whole, OSX-b, and SMPLer-X-b under three training settings, namely \textbf{P}, \textbf{Ours}, and \textbf{P+Ours}. Training exclusively on Human4K (\textbf{Ours}) leads to slightly higher body-related errors compared with \textbf{P}, which is expected because Human4K alone does not cover all appearance patterns and scene configurations present in the public datasets, yet it already yields improved hand-related metrics, indicating that the mocap-based extremity annotations in Human4K provide more precise supervision for hands.

When Human4K is combined with the public datasets (\textbf{P+Ours}), all three methods consistently deliver the best overall performance across benchmarks and body regions. On EHF, for instance, Hand4Whole reduces the overall MPVPE from 79.00 mm to 70.10 mm and the hand MPVPE from 44.11 mm to 40.12 mm after incorporating Human4K, indicating clear improvements in both global and fine-grained body estimation. OSX-b exhibits similar performance gains, with its overall MPVPE decreasing from 81.29 mm to 72.31 mm and hand MPVPE dropping from 60.43 mm to 48.98 mm. SMPLer-X-b also benefits substantially from Human4K, particularly in PA-MPVPE and facial regions, where the enhanced annotations lead to more accurate global alignment and refined local detail. On 3DPW, all three approaches achieve lower MPJPE and PA-MPJPE under the \textbf{P+Ours} setting compared to \textbf{P}, demonstrating stronger generalization to complex in-the-wild human poses.

Taken together, these results show that Human4K provides complementary high-fidelity supervision that not only strengthens extremity reconstruction, especially for hands, but also improves overall whole-body accuracy on standard benchmarks when used in conjunction with existing public datasets.

\begin{table}[!t]
\centering
\setlength{\tabcolsep}{6.5pt}
\renewcommand{\arraystretch}{1.2}
\caption{Comparative reconstruction errors on the Human4K test set with frontal view.}
\begin{tabular}{lccccc}
\toprule
\multirow{2}{*}{\textbf{Method}} 
 & \multirow{2}{*}{\textbf{Dataset}}  
 & \multicolumn{2}{c}{PA-MPVPE (\,mm)$\downarrow$}  
 & \multicolumn{2}{c}{PA-MPJPE (\,mm)$\downarrow$} \\
 & & All & Hand & Body & Hand \\
\midrule
\multirow{3}{*}{Hand4Whole} 
 & P         & 85.84 & 12.77 & 98.29 & 12.98 \\
 & Ours      & 81.64 & 12.61 & 93.98 & 11.24 \\
 & P+Ours    & \textbf{71.53} & \textbf{10.55} & \textbf{80.74} & \textbf{9.53} \\
\midrule
\multirow{3}{*}{OSX-b}      
 & P         & 93.12 & 13.93 & 107.16 & 14.11 \\
 & Ours      & 65.86 & 11.79 & 86.29 & 11.92 \\
 & P+Ours    & \textbf{56.15} & \textbf{10.89} & \textbf{61.94} & \textbf{9.97} \\
\midrule
\multirow{3}{*}{SMPLer-X-b} 
 & P         & 78.21 & 42.14 & 86.38 & 37.91 \\
 & Ours      & 75.18 & 39.42 & 82.16 & 35.60 \\
 & P+Ours    & \textbf{72.65} & \textbf{36.83} & \textbf{80.11} & \textbf{34.41} \\
\bottomrule
\end{tabular}
\label{tab5}
\end{table}

\begin{figure}[!t]
\centering
\includegraphics[width=\linewidth]{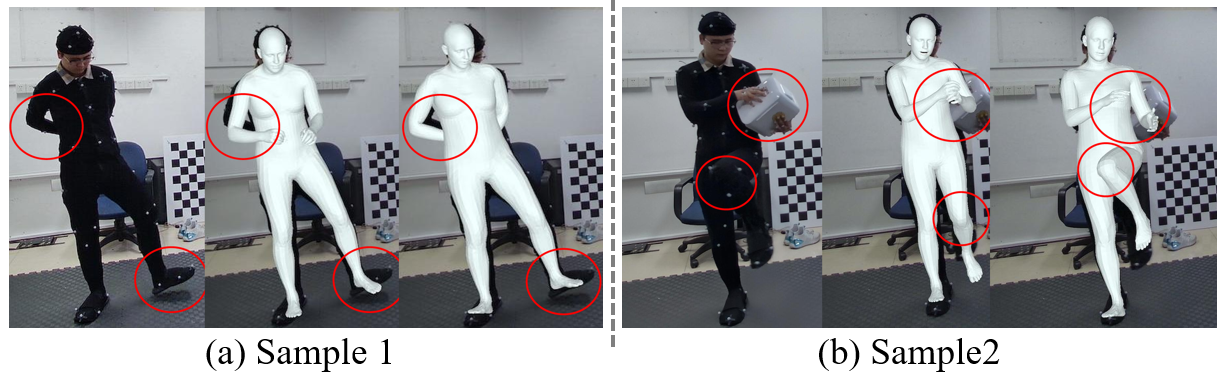}
\caption{\textbf{Effect of Human4K on challenging cases.}
For each sample, we show the RGB input, the reconstruction of models trained only on public datasets, and the reconstruction after adding Human4K. Human4K notably improves performance under depth ambiguity and strong limb occlusion.}
\label{fig7}
\end{figure}

\subsection{Establishing Human4K as an In-domain Benchmark}
We further evaluate the three baseline methods on the Human4K test set to examine how well existing models generalize to our high-resolution, highly articulated motion domain. As shown in Table~\ref{tab5}, models trained only on public datasets (P) perform poorly on Human4K, with large errors across both MPVPE and MPJPE metrics. This performance drop highlights that Human4K is substantially more challenging than standard benchmarks, due to its extreme limb articulations, frequent self-occlusion, and precise mocap-groundtruth supervision.

Training solely on Human4K (Ours) already yields notable improvements over P, indicating that public datasets do not cover the motion complexity present in Human4K. The best performance is achieved by jointly training on public datasets and Human4K (P+Ours). For example, OSX-b reduces PA-MPVPE (All) from 93.12\,mm to 56.15\,mm and PA-MPJPE (Body) from 107.16\,mm to 61.94\,mm after incorporating Human4K, corresponding to nearly 40\% relative improvement. Similar gains are observed for Hand4Whole and SMPLer-X-b. These results demonstrate that Human4K provides essential motion diversity and high-fidelity supervision that current models lack, and it serves as a strong in-domain benchmark where Human4K-based training is necessary for accurate full-body reconstruction.

Figure~\ref{fig7} further visualizes challenging cases with depth ambiguity and severe limb occlusion. Models trained only on traditional datasets frequently fail to recover correct limb orientation or body structure under these conditions, while models trained with additional Human4K produce more stable and accurate reconstructions. These qualitative observations align with the quantitative results, reinforcing that Human4K exposes failure modes not covered by existing datasets and provides the supervision needed to resolve them.

\begin{figure}[!t]
\centering
\includegraphics[width=0.85\linewidth]{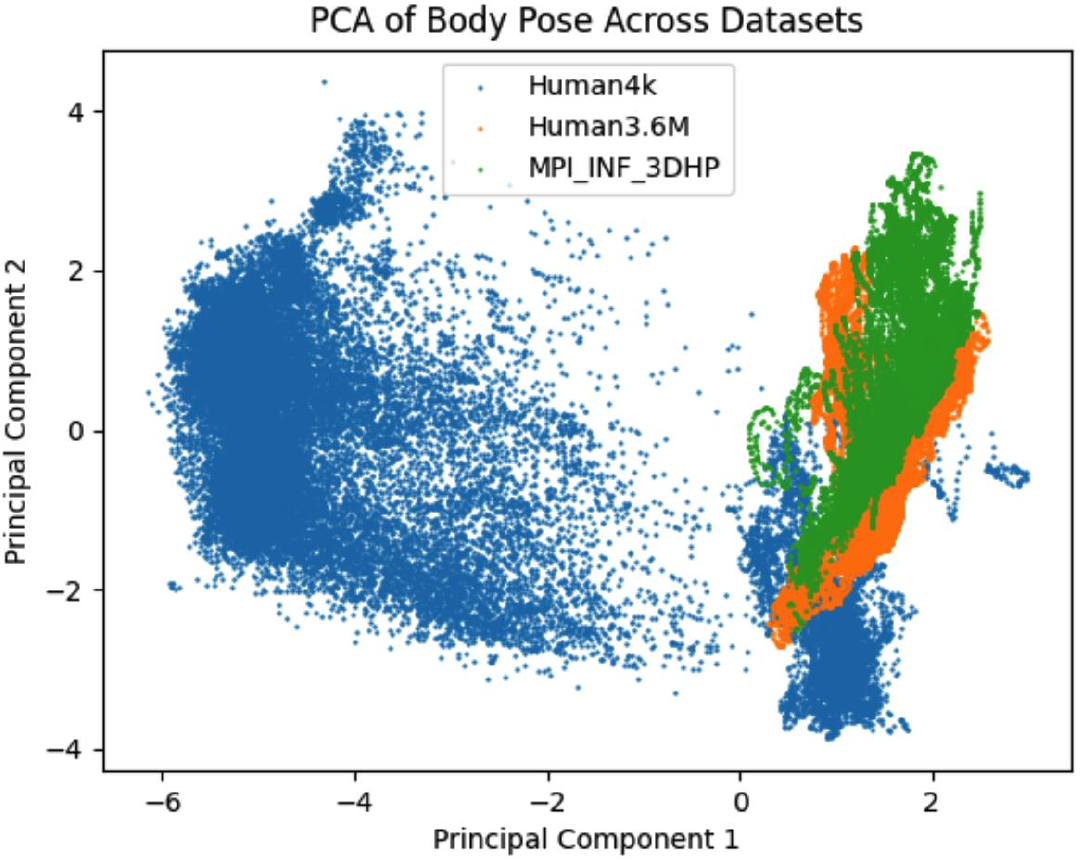}
\caption{The distribution of human body pose data from different datasets after dimensionality reduction using Principal Component Analysis (PCA)}
\label{fig8}
\end{figure}

\subsection{Characterizing Motion and Pose Diversity}
To better characterize the pose coverage of Human4K, we apply principal component analysis (PCA) to the 3D body joint coordinates from Human4K, Human3.6M, and MPI\_INF\_3DHP and visualize their distributions in the first two principal components, as shown in Fig.~\ref{fig8}. The samples from Human4K span a noticeably larger region in the PCA space, which indicates a wider spectrum of body configurations, motion patterns, and joint interactions. In contrast, the points from Human3.6M and MPI\_INF\_3DHP are concentrated in relatively compact areas, suggesting that their poses are more repetitive and constrained to fewer action types.

The clear separation between Human4K and the other two datasets implies that Human4K contains more complex postures, larger joint ranges, and more frequent self occlusions. This distributional difference aligns with our design goal that Human4K should provide richer and more challenging whole body motions. As a result, Human4K complements existing datasets by extending the pose space they cover and offers stronger supervision for models that aim to generalize to diverse and realistic 3D human motions.

\setlength{\tabcolsep}{14pt} 
\renewcommand{\arraystretch}{1.2}
\begin{table}[!t]
\centering
\caption{Comparison of 3D reconstruction errors for OSX-b under different image resolutions. Higher input resolutions consistently improve MPVPE, PA-MPVPE, and PA-MPJPE, especially for fine-grained hand and face regions.}
\begin{tabular}{@{}lccc@{}}
\toprule
\textbf{OSX-b} & \textbf{1K} & \textbf{2K} & \textbf{4K} \\
\midrule
\rowcolor{gray!15}
\multicolumn{4}{c}{\textbf{MPVPE (\,mm)$\downarrow$}} \\
All   & 107.71 & 100.63 & \textbf{94.19} \\
Hand  &  95.34 &  83.60 & \textbf{61.58} \\
Face  &  97.86 &  83.97 & \textbf{45.96} \\
\midrule
\rowcolor{gray!15}
\multicolumn{4}{c}{\textbf{PA-MPVPE (\,mm)$\downarrow$}} \\
All   &  102.52 & 94.64 & \textbf{87.09} \\
Hand  &  12.55 & 12.33 & \textbf{11.81} \\
Face  &   6.14 &  3.99 & \textbf{3.76} \\
\midrule
\rowcolor{gray!15}
\multicolumn{4}{c}{\textbf{PA-MPJPE (\,mm)$\downarrow$}} \\
Body  & 109.69 & 99.99 & \textbf{81.81} \\
Hand  &  11.52 & 10.68 & \textbf{10.11} \\
\bottomrule
\end{tabular}
\label{tab6}
% \vspace{-0.3cm}
\end{table}

\subsection{Understanding the Impact of 4K Resolution}
To quantify the impact of input resolution, we evaluate OSX-b on Human4K under three image resolutions, namely 1K, 2K, and 4K. The corresponding MPVPE, PA-MPVPE, and PA-MPJPE for the whole body, hands, and face are reported in Table~\ref{tab6}. As the input resolution increases, all error metrics consistently decrease across regions, indicating that richer spatial detail directly translates into more accurate 3D reconstruction. The gains are especially prominent for fine structures. Hand MPVPE is reduced from 95.34\,mm at 1K to 61.58\,mm at 4K, and face MPVPE decreases from 97.86\,mm at 1K to 45.96\,mm at 4K. For the body, PA-MPJPE also improves notably, dropping from 109.69\,mm at 1K to 81.81\,mm at 4K, which reflects a clear benefit even for larger-scale pose estimation.

These results demonstrate that high-resolution inputs are particularly important for modeling extremities and facial details, where subtle geometric cues are easily lost at lower resolutions. In practice, low resolution limits the visibility of fine edge contours, self occlusions, and small joint displacements, which leads to larger reconstruction errors, especially for hands and face. By contrast, 4K imagery preserves these high-frequency details and provides stronger supervision for both global body pose and local articulation. This experiment therefore provides quantitative support for our design choice to capture Human4K entirely in native 4K resolution, and highlights the value of high-resolution datasets for advancing high-precision whole-body 3D human reconstruction.

\begin{figure}[!hbpt]
\centering
\includegraphics[width=0.85\linewidth]{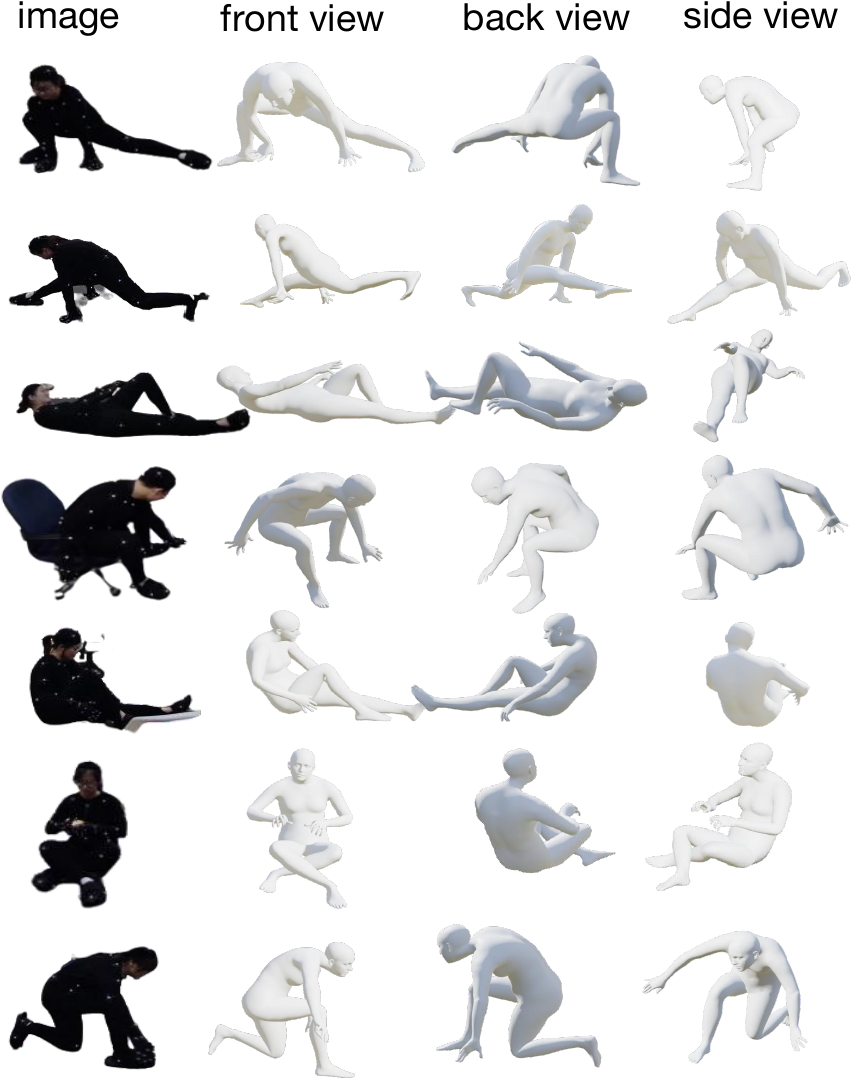} 
\caption{A sequence of human body poses captured from different angles (front, back, and side views) and depicted through 3D skeletal models generated from motion capture data.}
\label{fig9}
% \vspace{-0.3cm}
\end{figure}

\subsection{Qualitative Results}
Finally, we present qualitative examples to further illustrate the difficulty and annotation quality of Human4K. Figure~\ref{fig9} shows multi-view samples covering a wide range of action types, including sitting, bending, stretching, and sports movements, many of which involve strong self-occlusion and depth ambiguity. Across these challenging poses, the SMPL-X reconstructions remain well aligned with both the mocap skeleton and the image evidence from different viewpoints, demonstrating not only the motion diversity and complexity captured in Human4K but also the high precision and robustness of our mocap-based SMPL-X annotations.

\section{Conclusion}
In this work, we presented \textbf{Human4K}, a large-scale 4K multi-view motion capture dataset for high-precision whole-body 3D human reconstruction. Human4K is constructed through a unified acquisition and annotation pipeline that integrates synchronized 4K RGB imagery with Vicon mocap data, followed by SMPL-X motion retargeting and subsequent refinement of hand and face regions. The dataset contains over six million 4K frames from eight viewpoints and covers depth-ambiguous poses, strong self-occlusions and highly articulated limb motions. Training three representative SMPL-X based methods on Human4K leads to consistent gains on different benchmarks, while analyses of pose distribution and image resolution show that Human4K provides richer whole-body motion coverage and that native 4K input is especially beneficial for reconstructing extremity details, establishing Human4K as a strong real-world complement to current datasets.

\noindent\textbf{Limitations and Future Work.}
Although Human4K offers precise SMPL-X annotations, it is still captured in a controlled studio with actors wearing mocap suits, which reduces the diversity compared with real in-the-wild data. In future work, we will enrich appearance variability by recording with more varied scenes, and explore diffusion-based generative models to synthesize realistic changes in background and lighting so as to further improve the representativeness and application range of Human4K.

% \newpage
\printbibliography

\end{document}